\documentclass{article}

\usepackage{microtype}
\usepackage{graphicx}
\usepackage{subcaption}
\usepackage{booktabs} % for professional tables

\usepackage{hyperref}

\usepackage[preprint]{icml2026}

\usepackage{amsmath}
\usepackage{amssymb}
\usepackage{mathtools}
\usepackage{amsthm}

\usepackage[capitalize,noabbrev]{cleveref}

\theoremstyle{plain}

\theoremstyle{definition}

\theoremstyle{remark}

\usepackage{graphicx}
\usepackage{booktabs} % For professional-looking tables
\usepackage{subcaption}
\usepackage{multirow}
\usepackage{adjustbox} % For adjusting table width

\usepackage{xcolor}
\usepackage{colortbl}

\usepackage{fontawesome}
\usepackage{pifont}

\usepackage{wrapfig}

\newcommand{\model}{WISE} 

\newcommand{\onmark}{\textcolor{black}{\ding{51}}}
\newcommand{\offmark}{\textcolor{red!70!black}{}}

\usepackage[textsize=tiny]{todonotes}

\icmltitlerunning{Efficient Reasoning via Thought Compression for Language Segmentation}

\begin{document}

\twocolumn[
  \icmltitle{Efficient Reasoning via Thought Compression for Language Segmentation}



  \icmlsetsymbol{equal}{*}

  \begin{icmlauthorlist}
    \icmlauthor{Qing Zhou}{sch}
    \icmlauthor{Shiyu Zhang}{sch}
    \icmlauthor{Yuyu Jia}{sch}
    \icmlauthor{Junyu Gao}{sch}
    \icmlauthor{Weiping Ni}{comp}
    \icmlauthor{Junzheng Wu}{comp}
    \icmlauthor{Qi Wang}{sch}
  \end{icmlauthorlist}

  % \icmlaffiliation{sch}{Department of XXX, University of sch, Location, Country}
  \icmlaffiliation{sch}{Northwestern Polytechnical University, China}
  \icmlaffiliation{comp}{Northwest Institute of Nuclear Technology, China}

  \icmlcorrespondingauthor{Qing Zhou}{mrazhou@foxmail.com}
  \icmlcorrespondingauthor{Qi Wang}{crabwq@gmail.com}

  \icmlkeywords{Machine Learning, ICML}

% {  
%   \vskip 0.1in  % 控制图与上方 的距离
%   \centering
%   \includegraphics[width=0.98\textwidth]{figure/mov_v5.pdf}

%   \captionof{figure}{
%     \textbf{Left:} Our $\sqrt{r}$-scaling prevents the \textit{catastrophic accuracy collapse} of baselines (Vanilla/PA), recovering \textbf{+29\%} accuracy at aggressive compression.
%     \textbf{Middle:} In contrast to the artificial \textit{radial drift} in PA, our method prevents this inflation.
%     \textbf{Right:} Our approach restores angular fidelity (higher cosine similarity), effectively preserving feature semantics against heterogeneous merging errors.

  \vskip 0.25in
]



\printAffiliationsAndNotice{}  % no special notice (required even if empty)

\newcommand{\softmax}{\mathrm{softmax}}

\begin{abstract}
    Chain-of-thought (CoT) reasoning has significantly improved the performance of large multimodal models in language-guided segmentation, yet its prohibitive computational cost, stemming from generating verbose rationales, limits real-world applicability. We introduce WISE (Wisdom from Internal Self-Exploration), a novel paradigm for efficient reasoning guided by the principle of \textit{thinking twice---once for learning, once for speed}. WISE trains a model to generate a structured sequence: a concise rationale, the final answer, and then a detailed explanation. By placing the concise rationale first, our method leverages autoregressive conditioning to enforce that the concise rationale acts as a sufficient summary for generating the detailed explanation. This structure is reinforced by a self-distillation objective that jointly rewards semantic fidelity and conciseness, compelling the model to internalize its detailed reasoning into a compact form. At inference, the detailed explanation is omitted. To address the resulting conditional distribution shift, our inference strategy, WISE-S, employs a simple prompting technique that injects a brevity-focused instruction into the user's query. This final adjustment facilitates the robust activation of the learned concise policy, unlocking the full benefits of our framework. Extensive experiments show that WISE-S achieves state-of-the-art zero-shot performance on the ReasonSeg benchmark with 58.3 cIoU, while reducing the average reasoning length by nearly \textbf{5$\times$}---from 112 to just 23 tokens. Code is available at \href{https://github.com/mrazhou/WISE}{WISE}.
    \end{abstract}

\section{Introduction}
\label{sec:intro}

\begin{figure}[t]
    \centering
    \vspace{-15pt}
    \includegraphics[width=0.5\textwidth]{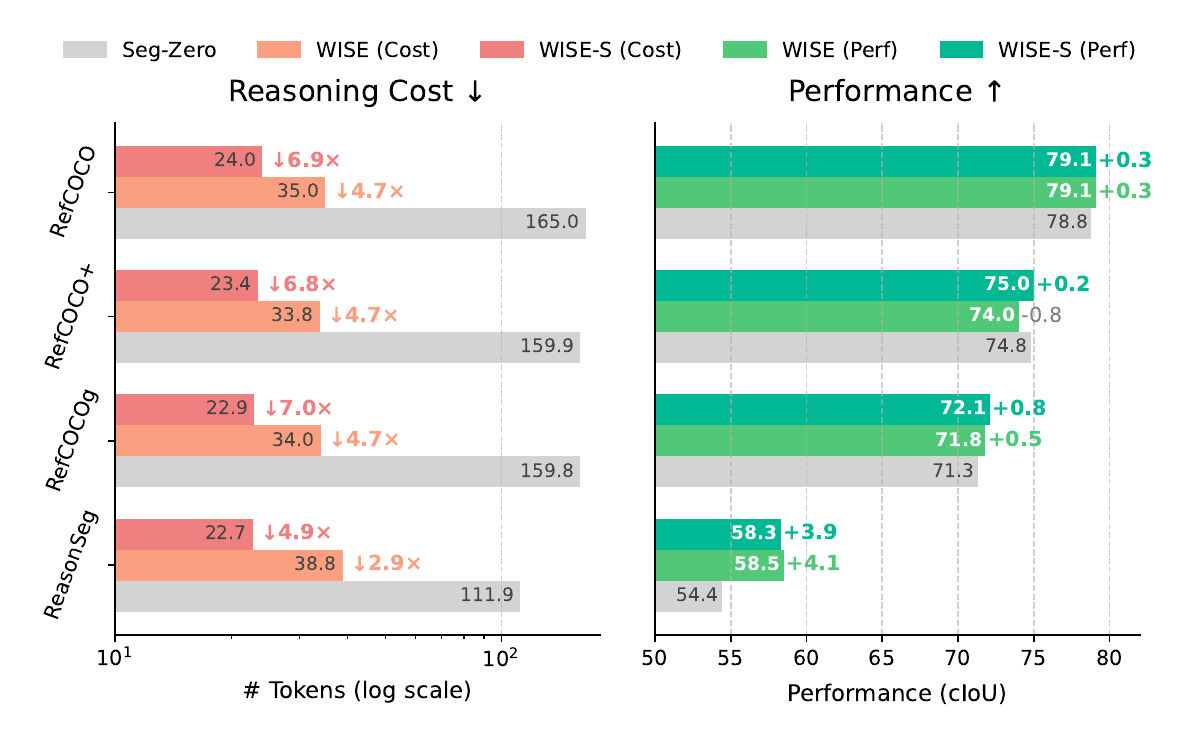}
    \captionof{figure}{
        % WISE balances cost and performance by separating reasoning for training and inference.
        \textbf{WISE achieves a superior cost-performance trade-off by decoupling reasoning for learning and inference.}
        Our framework trains the model with detailed explanations but uses a distilled, concise rationale at inference. 
        The base model, WISE, reduces reasoning cost by an average of 4.7$\times$ while outperforming the baseline. 
        Our final model, WISE-S, applies an additional inference-time prompt, achieving even greater compression (up to \textbf{10.6$\times$} on RefCOCO) and the highest overall performance. 
        On ReasonSeg, WISE-S simultaneously improves performance by \textbf{3.9 cIoU} and cuts token cost by \textbf{4.9$\times$}.
    }
    \label{fig:wise_cost_perf}
    \vspace{-10pt}
\end{figure}

Language-guided segmentation, the task of localizing an object within an image based on a natural language description, is a cornerstone of modern vision-language research. While early successes focused on simple referring expressions like ``the car'' \citep{yu2016refcoco,kazemzadeh2014referitgame}, the field's frontier has rapidly advanced towards \textit{reasoning segmentation} \citep{lai2024lisa}. This more challenging task requires models to interpret complex, multi-step commands that are often compositional, relational, or rely on commonsense knowledge---for instance, ``segment the car that is partially obscured by the bus.'' Successfully tackling such prompts necessitates a shift from basic recognition to a multi-step cognitive process, where a model must deconstruct the query into a coherent chain of visual search and logical validation steps.

Recent Large Multimodal Models (LMMs)~\citep{shang2024llava,bai2025qwen25}, augmented with Chain-of-Thought (CoT) prompting strategies~\citep{wei2022chain,ma2025cot}, have emerged as a powerful solution for this task. By generating an explicit, step-by-step textual rationale, these models can effectively reason through complex instructions to identify the correct target. However, this enhanced reasoning capability is intrinsically tied to the generation of long, verbose text sequences~\citep{wang2025pixelthink, feng2025efficient}. This creates a critical performance-efficiency paradox: the very mechanism that drives accuracy introduces prohibitive computational costs and high inference latency, severely limiting the applicability of these models in real-world scenarios like robotics or interactive assistants where efficiency is paramount. \textbf{The fundamental challenge, therefore, is to decouple a model's reasoning capability from this costly, verbose output.}

In this work, we address this challenge by proposing \textbf{WISE (Wisdom from Internal Self-Exploration)}, a novel training and inference paradigm designed for efficient reasoning via thought compression. Our framework is guided by the principle of \textit{thinking twice---once for learning, once for speed}. Instead of a simple think $\rightarrow$ answer sequence, we train the model to generate a structured, three-part response: a \textbf{concise rationale} ($R_c$), the \textbf{final answer} ($A$), and then a \textbf{detailed explanation} ($R_d$). By compelling the model to commit to the concise rationale upfront, we leverage the fundamental autoregressive nature of decoders. This structure ensures that the generation of the final answer and, critically, the detailed explanation are both conditioned on the initial concise thought, $P(A, R_d | I, T, R_c)$. This conditioning forces the concise rationale to become a potent and sufficient summary of the entire reasoning process, as a poor summary would make it impossible for the model to generate a coherent, logically consistent detailed explanation.

This unique training structure is reinforced by a self-distillation objective that explicitly rewards the semantic fidelity between the concise rationale and the detailed explanation, while penalizing the verbosity of the former. This process encourages the model to internalize its elaborate reasoning capabilities into a compact, efficient policy. To ensure this learned policy is robustly activated at inference, where the detailed explanation is entirely omitted to maximize speed, our WISE framework culminates in WISE-S, a simple, zero-overhead prompting strategy. This final adjustment injects a brevity-focused instruction into the user's query (as shown in the Fig.~\ref{fig:wise_cost_perf}), mitigating the conditional distribution shift between training and inference and ensuring the model consistently defaults to its more efficient reasoning mode.

Our contributions are as follows:
\begin{itemize}
    \item We introduce WISE, a end-to-end framework for efficient reasoning segmentation. It decouples the verbose reasoning required for robust learning from the concise rationale needed for fast inference.
    \item We propose a self-distillation mechanism, built upon a unique \textit{concise $\rightarrow$ answer $\rightarrow$ detailed} generation sequence that leverages autoregressive conditioning to effectively compress its reasoning process.
    \item We demonstrate through extensive experiments that our full framework, WISE-S, achieves state-of-the-art zero-shot performance on the ReasonSeg benchmark with 58.3 cIoU, while simultaneously reducing the average reasoning length by over \textbf{5$\times$}.
\end{itemize}

\section{Related Work}
\label{sec:related_work}

\subsection{Language-Guided Segmentation}
Language-guided segmentation aims to localize specific objects in an image based on natural language descriptions. Early work in this domain primarily focused on referring expression segmentation \citep{kazemzadeh2014referitgame,yu2016refcoco}, where queries are typically simple and direct, such as ``the person in the red shirt''. These foundational works established the core challenge of grounding linguistic phrases to pixel-level visual evidence.

The field has recently evolved towards the more challenging task of reasoning segmentation~\citep{lai2024lisa}. This advanced task requires models to interpret complex, multi-step instructions that involve relational, spatial, or commonsense reasoning. Several recent works have explored using LMMs to tackle this challenge, bridging the gap between high-level reasoning and low-level segmentation \citep{lai2024lisa, chen2024sam4mllm, ren2024pixellm}. A key work that pushed the boundaries of this task is Seg-Zero \citep{liu2025segzero}, which introduced a framework for learning reasoning capabilities "from zero". By employing pure reinforcement learning (RL) instead of supervised fine-tuning (SFT), Seg-Zero demonstrated that a model could learn to generate an explicit reasoning process and achieve strong zero-shot performance on complex benchmarks like ReasonSeg \citep{lai2024lisa}. Our work builds upon this paradigm, adopting the core idea of using a reasoning model to guide a segmentation model. However, we identify and address a critical limitation in these prior works: the high computational cost associated with their verbose reasoning chains.

\subsection{Chain-of-Thought in Large Multimodal Models}
Chain-of-Thought (CoT) prompting has emerged as a powerful technique for unlocking the complex reasoning abilities of Large Language Models (LLMs) \citep{wei2022chain,zhang2022automatic,yu2023towards}. By prompting a model to generate a sequence of intermediate reasoning steps before providing a final answer, CoT has significantly improved performance on a wide range of tasks, from arithmetic to symbolic reasoning. This principle has been successfully extended to LMMs, enabling them to deconstruct complex visual-linguistic queries into manageable steps \citep{chen2024sam4mllm,shen2025vlm,tan2025reason}.

In the context of reasoning segmentation, models like LISA \citep{lai2024lisa} and Seg-Zero \citep{liu2025segzero} implicitly leverage a CoT-like process by generating a textual rationale to guide their final segmentation decision. The success of these methods demonstrates the power of explicit reasoning for complex visual grounding. While highly effective, this approach creates a performance-efficiency paradox: the model's accuracy is directly coupled with the generation of long, verbose textual outputs, leading to high latency and computational overhead. This trade-off presents a major obstacle for real-world applications. Our work confronts this challenge by seeking to decouple reasoning capability from the generation of costly, verbose text.

\subsection{Knowledge Distillation and Thought Compression}
Knowledge distillation is a common technique for model compression, where a smaller "student" model is trained to mimic the output of a larger, more powerful "teacher" model. This concept has been adapted to distill complex reasoning processes from large models, often by training a smaller model on the rationales generated by a teacher \citep{lightman2023let, uesato2022solving}. Our work introduces a novel form of \textit{self-distillation} specifically designed for compressing the CoT process within a single model.

Unlike traditional distillation, which relies on a separate teacher model, WISE trains the model to be its own teacher. During training, the model generates both a concise rationale ($R_c$) and a detailed explanation ($R_d$). The core innovation lies in our training objective, which simultaneously rewards the semantic fidelity between the detailed "teacher" explanation and the concise "student" rationale, while penalizing the verbosity of the latter. The structured generation sequence ($R_c \rightarrow A \rightarrow R_d$) and autoregressive conditioning are critical mechanisms that enable this effective self-distillation. This approach fundamentally differs from prior methods like Seg-Zero, which learn to generate a single, often verbose, rationale through reinforcement learning \citep{guo2025deepseekr1} but lack an explicit mechanism for compression. WISE internalize and compress its reasoning process for efficient yet powerful reasoning segmentation.

\section{Methodology}
\label{sec:method}

\begin{figure*}[t]
    \centering
    \includegraphics[width=0.95\textwidth]{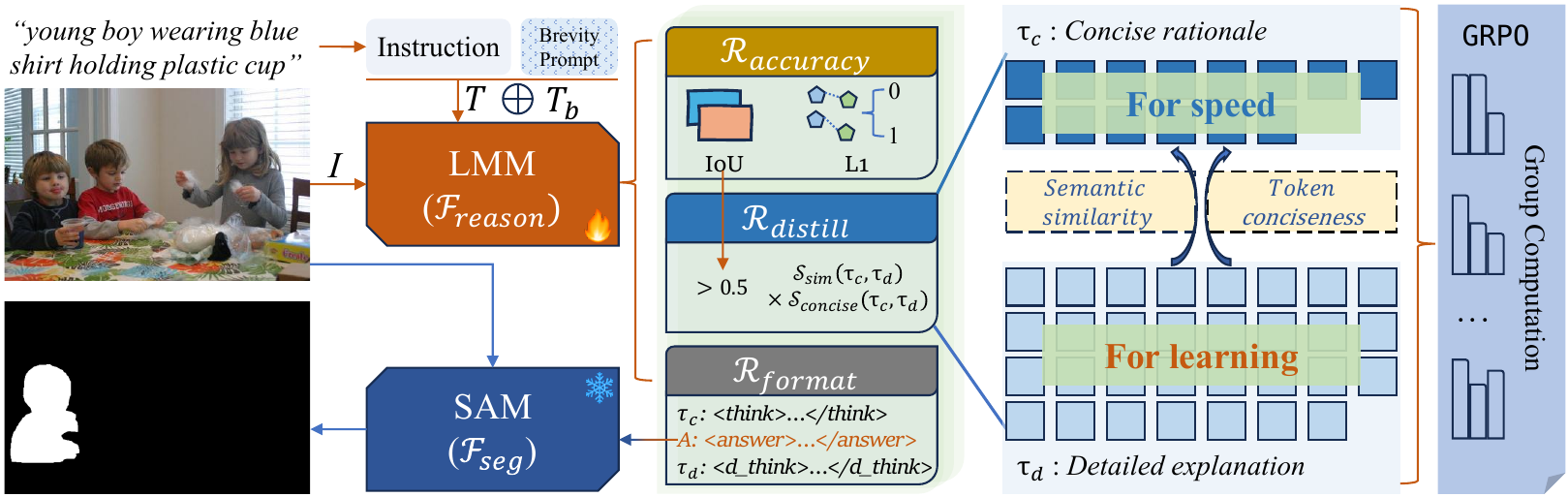}
    \caption{
        \textbf{Overview of the WISE framework.} During training (orange), the reasoning model $\mathcal{F}_{\text{reason}}$ generates a structured sequence: concise rationale ($\tau_c$), answer ($A$), and detailed explanation ($\tau_d$). The process is optimized via GRPO using a hierarchical reward, notably including a self-distillation term ($\mathcal{R}_{\text{distill}}$) that aligns $\tau_c$ with $\tau_d$ while enforcing brevity. The segmentation model $\mathcal{F}_{\text{seg}}$ remains frozen. At inference (blue), $\tau_d$ is omitted and a brevity prompt ($T_b$) is added to ensure highly efficient reasoning.
    }
    \label{fig:wise}
\end{figure*}

Our objective is to enhance the efficiency of reasoning segmentation models by compressing their Chain-of-Thought (CoT) process, without sacrificing performance. As shown in \ref{fig:wise}, we introduce \textbf{WISE}, a new training and inference paradigm that redefines the learning objective for a policy-based reasoning model, $\mathcal{F}_{\text{reason}}$. This model is trained using the GRPO reinforcement learning algorithm~\citep{shao2024deepseekmath} and is decoupled from the downstream segmentation model, $\mathcal{F}_{\text{seg}}$, which remains frozen throughout the process.

\subsection{Problem Formulation}
\label{ssec:problem_formulation}
Given an input image $I$ and a textual instruction $T$, our goal is to train a policy $\pi_{\theta}$, embodied by the reasoning model $\mathcal{F}_{\text{reason}}$, to generate an optimal set of geometric prompts $A$ (e.g., bounding boxes and points) that localize the target object. The baseline approach trains the policy to generate a sequence containing a detailed reasoning chain $\tau_d$ and the prompts $A$. The optimization objective is to find parameters $\theta^*$ that maximize the expected reward, which is calculated directly on the generated prompts:
\begin{equation}
    \theta^* = \arg\max_{\theta} \mathbb{E}_{\pi_{\theta}}[\mathcal{R}_{\text{task}}(A, A_{gt})]
\end{equation}
where $A_{gt}$ represents the ground-truth geometric prompts. For final evaluation, the generated prompts $A$ are fed into a fixed, pre-trained segmentation model $M = \mathcal{F}_{\text{seg}}(A)$, but $\mathcal{F}_{\text{seg}}$ and the final mask $M$ are not involved in the training loop of $\pi_{\theta}$. The primary motivation for our work is the high computational cost of generating the verbose $\tau_d$ at inference time.

\subsection{The WISE Training Paradigm}
\label{ssec:wise_training}
The core of WISE is to train the policy $\pi_{\theta}$ to learn a compressed reasoning representation. This is achieved by restructuring the generation sequence as a structured action space and introducing a hierarchical, conditional reward objective.

\paragraph{Structured Action Space and Autoregressive Conditioning}
\label{sssec:autoregressive}
Instead of the standard action space $\mathcal{A}_{\text{baseline}} = (\tau_d, A)$, we define a new action space for training: $\mathcal{A}_{\text{train}} = (\tau_c, A, \tau_d)$, where $\tau_c$ is a concise rationale (\texttt{<think>}), $A$ is the answer (\texttt{<answer>}), and $\tau_d$ is a detailed explanation (\texttt{<d\_think>}). The autoregressive policy $\pi_{\theta}$ models the joint probability of this sequence as:
\begin{equation}
\label{eq:autoregressive}
\begin{aligned}
    \pi_{\theta}(\tau_c, A, \tau_d | I, T) = & \pi_{\theta}(\tau_c | I, T) \\
    & \cdot \pi_{\theta}(A | I, T, \tau_c) \\
    & \cdot \pi_{\theta}(\tau_d | I, T, \tau_c, A)
\end{aligned}
\end{equation}
% \begin{equation}
% \label{eq:autoregressive}
% \begin{aligned}
%     \pi_{\theta}(\tau_c, A, \tau_d | I, T) = & \pi_{\theta}(\tau_c | I, T) \cdot \pi_{\theta}(A | I, T, \tau_c) \cdot \pi_{\theta}(\tau_d | I, T, \tau_c, A)
% \end{aligned}
% \end{equation}
This generation order imposes a powerful structural prior, compelling $\tau_c$ to act as a sufficient statistic for the generation of $\tau_d$, thereby creating ideal conditions for self-distillation.

\paragraph{Hierarchical Reward Objective}
\label{sssec:reward}
The learning process is guided by a hierarchical reward function. The base task reward, $\mathcal{R}_{\text{task}}$, ensures syntactic and geometric fidelity. It is a composite function composed of a format reward, $\mathcal{R}_{\text{format}}$, and an accuracy reward, $\mathcal{R}_{\text{accuracy}}$:
\begin{equation}
    \mathcal{R}_{\text{task}}(\tau_c, A) = \mathcal{R}_{\text{format}}(\tau_c, A) + \mathcal{R}_{\text{accuracy}}(A, A_{gt})
\end{equation}
Following prior work \citep{liu2025segzero}, $\mathcal{R}_{\text{format}}$ is a binary reward verifying the presence of required tags and a valid JSON structure. $\mathcal{R}_{\text{accuracy}}$ is the sum of binary rewards for the Intersection over Union (IoU) and L1 distance of the geometric prompts in $A$.

Building upon this baseline, the total training reward, $\mathcal{R}_{\text{train}}$, conditionally incorporates our novel self-distillation reward, $\mathcal{R}_{\text{distill}}$:
\begin{equation}
\label{eq:total_reward}
\mathcal{R}_{\text{train}} = \mathcal{R}_{\text{task}} + \mathcal{R}_{\text{distill}}(\tau_c, \tau_d) \cdot \mathbb{I}(\text{IoU}(A) > 0.5)
\end{equation}
where $\mathbb{I}(\cdot)$ is the indicator function. The distillation reward $\mathcal{R}_{\text{distill}}$ is defined as the product of a similarity score and a conciseness score:
\begin{equation}
\label{eq:distill_reward}
\mathcal{R}_{\text{distill}} = \mathcal{S}_{\text{sim}}(\tau_c, \tau_d) \cdot \mathcal{S}_{\text{concise}}(\tau_c, \tau_d)
\end{equation}
The semantic similarity score, $\mathcal{S}_{\text{sim}}$, is computed as the cosine similarity between the sentence embeddings of the two rationales, obtained from a pretrained SentenceTransformer model $\mathcal{F}_{\text{ST}}$:
\begin{equation}
\label{eq:sim_score}
\mathcal{S}_{\text{sim}}(\tau_c, \tau_d) = \frac{\mathcal{F}_{\text{ST}}(\tau_c) \cdot \mathcal{F}_{\text{ST}}(\tau_d)}{\|\mathcal{F}_{\text{ST}}(\tau_c)\| \|\mathcal{F}_{\text{ST}}(\tau_d)\|}
\end{equation}
The conciseness score, $\mathcal{S}_{\text{concise}}$, measures the fractional length reduction of the concise rationale relative to the detailed one, given by:
\begin{equation}
\label{eq:concise_score}
\mathcal{S}_{\text{concise}}(\tau_c, \tau_d) = \max\left(0, 1 - \frac{\text{len}(\tau_c)}{\text{len}(\tau_d)}\right)
\end{equation}
where $\text{len}(\cdot)$ denotes the number of tokens. This conditional objective ensures the model only learns to distill reasoning pathways that are proven to be effective.

\begin{figure}
    \centering
    \includegraphics[width=0.47\textwidth]{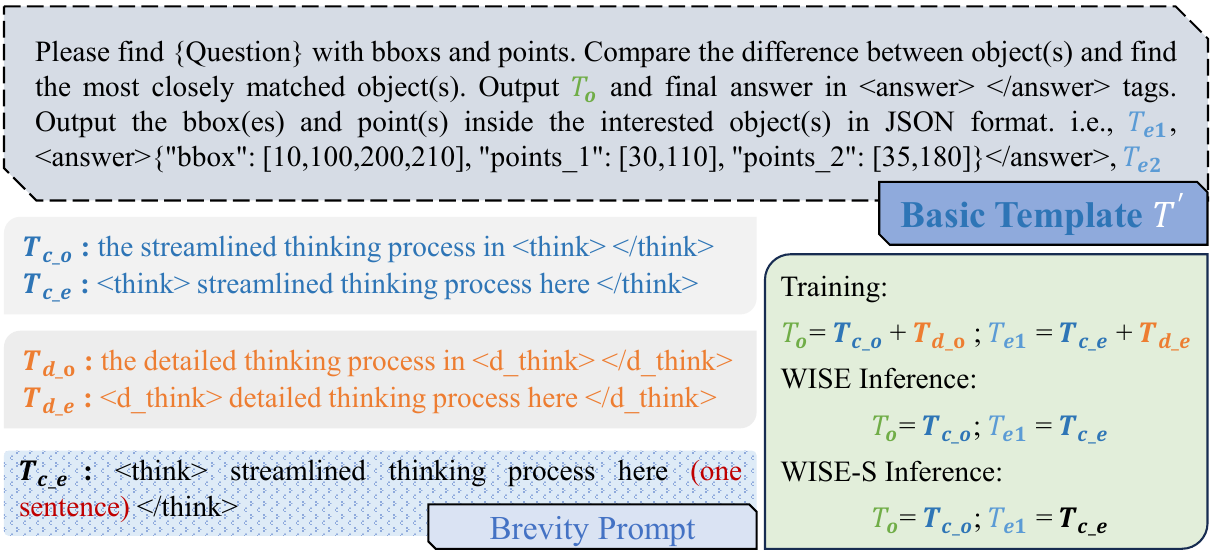}
    % \caption{User instruction for training and inference.}
    \caption{\textbf{Prompting Strategy for WISE.} The figure shows the basic instruction template and the specific components for generating concise ($T_c$) and detailed ($T_d$) rationales. Different combinations of these components are used for the training phase, standard WISE inference, and the brevity-focused WISE-S inference.}
    \label{fig:user_instruction}
\end{figure}

\subsection{The WISE-S Inference Strategy}
\label{ssec:wise_s_inference}
At inference time, the action space is reduced to $\mathcal{A}_{\text{infer}} = (\tau_c, A)$ by omitting the generation of $\tau_d$. This creates a conditional distribution shift, as the policy is now queried for $P(\tau_c | I, T; \text{goal}=A)$ instead of the training-time distribution that anticipated generating $\tau_d$. To mitigate this, the WISE-S strategy modifies the input instruction $T$ with a brevity-focused prompt $T_b=\textit{`the shorter the better'}$, creating a new inference prompt $T_S = T \oplus T_b$, where $\oplus$ denotes injection.
The objective is to align the inference-time policy with the characteristics learned during training, specifically to promote the generation of a concise rationale ($\tau_c$) that is as brief as its training-time counterpart. This is reflected in the following approximation:
\begin{equation}
    \pi_{\theta}(\tau_c, A | I, T_S) \approx \pi_{\theta}(\tau_c, A | I, T; \text{trained with } \mathcal{R}_{\text{distill}})
\end{equation}
This final step ensures the model robustly produces the highly compressed rationales learned during the WISE training phase.

\section{Experiments}
\label{sec:experiments}

\noindent\textbf{Datasets and Metrics.}
To empirically validate our framework, we largely adhere to the evaluation protocols established by prior work~\citep{liu2025segzero}. We train our models on a 2,000-sample subset of the RefCOCOg dataset~\citep{yu2016refcoco}. Crucially, no human-annotated rationales are used during training, meaning the reasoning capabilities are learned from scratch. For evaluation, we assess out-of-domain reasoning capability on the challenging ReasonSeg benchmark~\citep{lai2024lisa} and test in-domain performance on the standard suites of RefCOCO, RefCOCO+, and RefCOCOg~\citep{yu2016refcoco}. Segmentation accuracy is measured primarily by cumulative IoU (cIoU), while reasoning efficiency is quantified by the average number of generated reasoning tokens.

\noindent\textbf{Baselines.}
Our primary baseline is Seg-Zero, which represents a strong, reasoning-augmented segmentation model. To demonstrate that WISE's effectiveness stems from its unique training structure rather than just a preference for shorter outputs, we implement two additional strong baselines~\citep{aggarwal2025l1} that directly optimize for brevity via reward shaping:
\begin{itemize}
    \item \textbf{L1-Exact:} This baseline adds a reward term that is inversely proportional to the L1 distance between the generated rationale's length and a fixed, short target length.
    \item \textbf{L1-Max:} This baseline introduces a penalty for any rationale exceeding a predefined maximum length, encouraging the model to stay within a token budget.
\end{itemize}

\noindent\textbf{Implementation Details.}
Our model architecture comprises a Qwen2.5-VL-7B~\citep{bai2025qwen25} reasoning model ($\mathcal{F}_{\text{reason}}$) and a frozen SAM2-Large~\citep{ravi2024sam2} segmentation model ($\mathcal{F}_{\text{seg}}$). We optimize the reasoning model using the GRPO reinforcement learning algorithm~\citep{shao2024deepseekmath} with the DeepSpeed library~\citep{rasley2020deepspeed}. Training is conducted with a total batch size of 16, a learning rate of $1 \times 10^{-6}$, and the AdamW optimizer with a weight decay of 0.01. The user instruction for training and inference is shown in Figure~\ref{fig:user_instruction}.

\noindent\textbf{Inference Strategies.}
We evaluate two variants of our trained model at inference time:
\begin{itemize}
    \item \textbf{WISE (default):} This is our standard inference setting used across all experiments unless otherwise specified. It simply involves removing the instruction to generate the detailed explanation (\texttt{<d\_think>}) from the training-time prompt, relying on the model's naturally learned concise policy.
    \item \textbf{WISE-S (shortened):} To fully and robustly activate the learned brevity, this variant incorporates an additional brevity-focused prompt, such as ``\textit{one sentence}'', into the user instruction.
\end{itemize}

\begin{table}[t]
    \centering
    \caption{Comparison with state-of-the-art methods on the \textbf{Zero-shot ReasonSeg} benchmark. WISE-7B and WISE-7B-S significantly outperform prior methods. $^{*}$We re-evaluate with their official checkpoints. $^{\dagger}$Results reported in the original paper using different checkpoints.}
    \small
    % \adjustbox{max width=\linewidth}{
    \setlength{\tabcolsep}{2pt}
    \begin{tabular}{l|cc|cc}
    \toprule[1.25pt]
    \rowcolor{gray!11} & \multicolumn{4}{c}{\textbf{ReasonSeg}} \\
    \rowcolor{gray!11} & \multicolumn{2}{c|}{val} & \multicolumn{2}{c}{test} \\
    \rowcolor{gray!11}\multicolumn{1}{c|}{\multirow{-3}{*}{Method}} & gIoU & cIoU & gIoU & cIoU \\
    \midrule\midrule
    OVSeg~\cite{liang2023ovseg} & 28.5 & 18.6 & 26.1 & 20.8 \\
    ReLA~\cite{liu2023rela}    & 22.4 & 19.9 & 21.3 & 22.0 \\
    Grounded-SAM~\cite{ren2024groundedsam} & 26.0 & 14.5 & 21.3 & 16.4 \\
    LISA-7B-LLaVA1.5~\cite{lai2024lisa} & 53.6 & 52.3 & 48.7 & 48.8 \\
    LISA-13B-LLaVA1.5~\cite{lai2024lisa} & 57.7 & 60.3 & 53.8 & 50.8 \\
    SAM4MLLM~\cite{chen2024sam4mllm}  & 46.7 & 48.1 & - & - \\
    Qwen2.5VL-3B + SAM2 & 53.8 & 44.1 & 47.6 & 37.4 \\
    \textcolor{gray!70}{Seg-Zero-7B$^{\dagger}$~\cite{liu2025seg}} & \textcolor{gray!70}{62.6} & \textcolor{gray!70}{62.0} & \textcolor{gray!70}{57.5} & \textcolor{gray!70}{52.0} \\ 
    \midrule
    Seg-Zero-7B$^{*}$~\cite{liu2025seg} & 60.9 & 57.4 & 57.7 & 54.4 \\ 
    \midrule
    \textcolor{black}{\textbf{\model-7B (ours)}} & \textbf{{63.5}} & \textbf{59.2} & \textbf{60.3} & \textbf{58.5} \\
    \textcolor{black}{\textbf{\model-7B-S (ours)}} & \textbf{63.5} & \underline{58.8} & \textbf{60.3} & \underline{58.3} \\
    \bottomrule
    \end{tabular}
    % }
    \label{tab:reasonseg_bench}
    \end{table}
    
\begin{table}[t]
    \centering
    % \adjustbox{max width=\linewidth}{
    \caption{Comparison with state-of-the-art methods on the \textbf{Referring Expression Segmentation} benchmark. Our models achieve competitive or superior performance against specialized methods while being more general.}
    \small
    \setlength{\tabcolsep}{3pt}
    \begin{tabular}{l|ccc}
    \toprule[1.25pt]
    \rowcolor{gray!11} & \multicolumn{1}{c}{\textbf{RCO}} & \multicolumn{1}{c}{\textbf{RCO+}} & \multicolumn{1}{c}{\textbf{RCOg}} \\
    \rowcolor{gray!11}\multicolumn{1}{c|}{\multirow{-2}{*}{Method}} & testA & testA & test  \\
    \midrule\midrule
    LAVT~\cite{yang2022lavt}        & 75.8 & 68.4 & 62.1  \\
    ReLA~\cite{liu2023rela}        & 76.5 & 71.0 & 66.0  \\
    LISA-7B~\cite{lai2024lisa}      & 76.5 & 67.4 & 68.5  \\
    PixelLM-7B~\cite{ren2024pixellm}     & 76.5 & 71.7 & 70.5  \\
    MagNet~\cite{chng2024mask} & 78.3 & 73.6 & 69.3  \\
    PerceptionGPT-7B~\cite{pi2024perceptiongpt}  & 78.6 & 73.9 & 71.7  \\
    % Seg-Zero-3B$^{\dagger}$~\cite{liu2025seg} & {79.3} &  {73.7}& {71.5} \\
    \textcolor{gray!70}{Seg-Zero-7B$^{\dagger}$~\cite{liu2025seg}} & \textcolor{gray!70}{80.3} & \textcolor{gray!70}{76.2} & \textcolor{gray!70}{72.6} \\ 
    \midrule

    Seg-Zero-7B$^{*}$~\cite{liu2025seg} & 78.8  & \underline{74.8} & 71.3  \\ 
    \midrule

    \textcolor{black}{\textbf{\model-7B (ours~)}} & \textbf{{79.1}} & 74.0 &  \underline{71.8} \\
    \textcolor{black}{\textbf{\model-7B-S (ours~)}} & \textbf{{79.1}} & \textbf{75.0} &  \textbf{72.1} \\

    \bottomrule
    \end{tabular}
    \label{tab:refcoco_bench}
\end{table}

\begin{table*}[t]
    \centering
    \caption{Performance and Efficiency Comparison on Referring and Reasoning Segmentation Benchmarks. Our WISE models drastically reduce token overhead (\#Tok) while improving or maintaining accuracy (cIoU) across all benchmarks compared to the Seg-Zero and brevity-focused reward shaping methods (L1-Exact, L1-Max~\citep{aggarwal2025l1}). The sub-rows show the token reduction factor ($\times\downarrow$) and the absolute performance change ($\Delta$) against the Seg-Zero.}
    \small
    \begin{tabular}{l|cccccc|cccc}
    \toprule
    % \rowcolor{gray!15}
    & \multicolumn{2}{c}{\textbf{RefCOCO$_{testA}$}} & \multicolumn{2}{c}{\textbf{RefCOCO+ $_{testA}$}} & \multicolumn{2}{c|}{\textbf{RefCOCOg $_{test}$}} & \multicolumn{4}{c}{\textbf{ReasonSeg (cIoU)}} \\
    % \rowcolor{gray!15}
    \multicolumn{1}{c|}{\multirow{-2}{*}{Method}} & \#Tok $\downarrow$ & cIoU $\uparrow$ & \#Tok $\downarrow$ & cIoU $\uparrow$ & \#Tok $\downarrow$ & cIoU $\uparrow$ & \#Tok $\downarrow$ & val $\uparrow$ & \#Tok $\downarrow$ & test $\uparrow$ \\
    \midrule \midrule
    
    Seg-Zero & 165.1 & 78.8 & 159.9 & 74.8 & 159.8 & 71.3 & 117.5 & 57.4 & 111.9 & 54.4 \\
    \midrule
    
    Seg-Zero+L1-Exact & 31.2 & 78.3 & 31.1 & 74.9 & 30.7 & 71.7 & 36.6 & 57.2 & 35.6 & 55.0 \\
    Seg-Zero+L1-Max & \textbf{11.7} & 78.9 & \textbf{11.7} & 74.6 & \textbf{10.6} & 70.7 & \textbf{10.5} & 43.8 & \textbf{12.0} & 49.7 \\
    
    \midrule
        
    \textbf{WISE} (ours) & 35.0 & \textbf{79.1} & 33.8 & 74.0 & 34.0 & 71.8 & 38.8 & \textbf{59.2} & 38.8 & \textbf{58.5} \\
    \rowcolor{gray!12}
    \textit{\small \quad vs. Seg-Zero} & \textit{\small (4.7$\times\downarrow$)} & \textit{\small (+0.3)} & \textit{\small (4.7$\times\downarrow$)} & \textit{\small (-0.8)} & \textit{\small (4.7$\times\downarrow$)} & \textit{\small (+0.5)} & \textit{\small (3.0$\times\downarrow$)} & \textit{\small (+1.8)} & \textit{\small (2.9$\times\downarrow$)} & \textit{\small (+4.1)} \\
    
    \midrule
    
    \textbf{WISE-S} (ours) & \underline{24.0} & \textbf{79.1} & \underline{23.4} & \textbf{75.0} & \underline{22.9} & \textbf{72.1} & \underline{24.6} & 58.8 & \underline{22.7} & 58.3 \\
    \rowcolor{gray!12}
    \textit{\small \quad vs. Seg-Zero} & \textit{\small (\textbf{6.9$\times\downarrow$})} & \textit{\small (+0.3)} & \textit{\small (\textbf{6.8$\times\downarrow$})} & \textit{\small (+0.2)} & \textit{\small (\textbf{7.0$\times\downarrow$})} & \textit{\small (+0.8)} & \textit{\small (\textbf{4.8$\times\downarrow$})} & \textit{\small (+1.4)} & \textit{\small (\textbf{4.9$\times\downarrow$})} & \textit{\small (+3.9)} \\
    
    \bottomrule
    \end{tabular}
    \label{tab:main_results_full_delta}
    \end{table*}

% Table R2: Generalization
\begin{table}[h]
    \centering
    \caption{\textbf{Generalization to Unified Tasks.} Validating WISE on the VisionReasoner across three distinct visual reasoning tasks. Note that VisionReasoner-7B-S applies our brevity prompt to the baseline, confirming the prompt's generality but highlighting the superior compression of the full WISE training.}
    \label{tab:generalization}
    \setlength{\tabcolsep}{3pt}
    \small
    \begin{tabular}{l|cc|cc|cc}
        \toprule
        \multirow{2}{*}{Method} & \multicolumn{2}{c|}{Det.$_{\texttt{COCO}}$} & \multicolumn{2}{c|}{Seg$_{\texttt{ReaSeg}}$} & \multicolumn{2}{c}{Count$_{\texttt{Pixmon}}$} \\
        & {\#Tok} & {mAP} & {\#Tok} & {gIoU} & {\#Tok} & {Acc} \\
        \midrule
        VisionReasoner-7B & 62.5 & 38.8 & 80.1 & 64.0 & 74.5 & 70.1 \\
        VisionReasoner-7B-S & 42.1 & 39.1 & 45.0 & 63.6 & 53.7 & 69.3 \\
        \midrule
        \textbf{+WISE (Ours)} & 67.7 & 39.5 & 63.0 & 66.0 & 84.9 & 68.2 \\
        \textbf{+WISE-S (Ours)} & \textbf{15.7} & \textbf{39.6} & \textbf{17.1} & \textbf{66.1} & \textbf{19.2} & \textbf{73.0} \\
        \bottomrule
    \end{tabular}
\end{table}

\subsection{Main Results}
\label{ssec:main_results}

We conduct a series of experiments to validate the effectiveness and efficiency of our proposed WISE framework. We first present the main results, comparing WISE with state-of-the-art methods on both reasoning and referring segmentation benchmarks. We then conduct extensive ablation studies to deconstruct the key components of our method and analyze their individual contributions.

\noindent\textbf{State-of-the-Art on Reasoning Segmentation.}
As shown in Table~\ref{tab:reasonseg_bench}, on the challenging zero-shot ReasonSeg benchmark, our WISE models significantly outperform existing state-of-the-art methods. Specifically, our default WISE-7B model achieves 60.3 gIoU and 58.5 cIoU on the test set, surpassing the strong LISA-13B baseline by a large margin (+6.5 gIoU, +7.7 cIoU). This demonstrates the strong reasoning capability cultivated by our training paradigm. The shortened variant, WISE-7B-S, maintains this high level of performance, confirming that the concise rationales preserve the essential logical steps required for complex reasoning.

\noindent\textbf{Competitive Performance on Referring Segmentation.}
Table~\ref{tab:refcoco_bench} shows the performance on referring expression segmentation benchmarks. Despite being trained on a small subset of RefCOCOg, our models exhibit strong generalization. WISE-7B-S achieves competitive or even superior results compared to specialized methods, particularly on RefCOCO+ and RefCOCOg, highlighting its robustness and wide applicability without sacrificing performance on simpler, in-domain tasks.

\noindent\textbf{Breaking the Efficiency-Performance Trade-off.}
Table~\ref{tab:main_results_full_delta} provides a comprehensive comparison of performance versus efficiency. Across all four benchmarks, both WISE and WISE-S drastically reduce the number of generated tokens compared to the Seg-Zero baseline. WISE-S, in particular, achieves a remarkable \textbf{4.9$\times$} to \textbf{7.0$\times$} reduction in reasoning length. More importantly, this massive gain in efficiency is not achieved at the cost of accuracy. In most cases, WISE-S \textit{improves} performance, especially on the difficult ReasonSeg task (+3.9 cIoU). This result empirically validates our core thesis: by teaching a model to reason efficiently, we can break the conventional trade-off and achieve both speed and effectiveness simultaneously.

\noindent\textbf{Generalization to Unified Visual Reasoning.} 
To demonstrate that WISE is task-agnostic, we extend our evaluation to VisionReasoner~\cite{liu2025visionreasoner}, a unified pipeline for Detection~\cite{lin2014microsoft}, Segmentation, and Counting~\cite{deitke2025molmo}. 
We also apply our prompting strategy to the baseline (VisionReasoner-7B-S) for a fair comparison.
As shown in Table~\ref{tab:generalization}, WISE-S demonstrates remarkable generalization.
Crucially, on the Detection task, the default WISE model generates \textit{more} tokens (67.7) than the baseline (62.5), revealing that the implicit policy lacks the robustness to maintain conciseness consistently across diverse task formats.
However, applying the WISE-S prompt effectively bridges this gap, aligning the inference behavior with the training objective. 
Most importantly, WISE-S achieves this extreme compression (e.g., reducing Detection tokens to {15.7}) while simultaneously {boosting performance}: Detection mAP improves to {39.6} (vs. 38.8 Base) and Counting Accuracy rises to {73.0} (vs. 70.1 Base).
This confirms that WISE-S is essential to robustly unlock the learned efficiency, translating concise reasoning into superior accuracy.

\subsection{Ablation Studies}
\label{sec:ablation}

To validate the design choices of WISE, we conduct component-wise ablations using the 7B model. We examine three critical aspects: generation order, reward composition, and inference-time prompting.

% \vspace{2pt}
\noindent\textbf{Impact of Generation Order.}
We first verify the core hypothesis of thought compression: that predicting the concise rationale \textit{before} the detailed one enables abstraction. As shown in Table~\ref{tab:abla_thought_order}, the proposed order $\tau_c \rightarrow A \rightarrow \tau_d$ is superior. Inverting the order ($\tau_d \rightarrow \tau_c$) fails to reduce token usage at inference (106.5 tokens), as the model conditions on the verbose chain rather than learning to abstract.

\noindent\textbf{Necessity of Reward Components.}
Table~\ref{tab:abla_reward_components} dissects the self-distillation reward. Removing the semantic similarity constraint ($S_{sim}$) degrades conciseness, as the model lacks the incentive to align $\tau_c$ with the detailed reasoning. Similarly, removing the conciseness penalty ($S_{concise}$) results in longer outputs. The conditional indicator ($\mathbb{I}$) proves crucial for stability; training without it (distilling failed reasoning paths) leads to performance degradation.

\begin{table}[h]
    \centering
    \caption{\textbf{Ablation on Generation Order.} Placing the concise rationale first ($\tau_c \rightarrow A \rightarrow \tau_d$) is essential for enabling effective thought compression at inference time.}
    \small
    \setlength{\tabcolsep}{4pt}
    \begin{tabular}{l|cc|cc}
        \toprule[1pt]
            \multirow{2}{*}{Order} & \multicolumn{2}{c|}{RefCOCOg} & \multicolumn{2}{c}{ReasonSeg} \\
            & \#Tok $\downarrow$ & cIoU $\uparrow$ & \#Tok $\downarrow$ & cIoU $\uparrow$ \\
            \midrule
            $A \rightarrow\tau_c\rightarrow\tau_d$ & 74.6 & 70.1 & 75.7 & 54.4 \\
            $\tau_d\rightarrow\tau_c\rightarrow A$ & 107.5 & 68.2 & 106.5 & 55.4 \\
            $\tau_c\rightarrow\tau_d\rightarrow A$ & 134.7 & 70.1 & 132.7 & 55.2 \\
            \rowcolor{gray!10} \textbf{$\tau_c\rightarrow A\rightarrow\tau_d$ (Ours)} & \textbf{34.0} & \textbf{71.8} & \textbf{38.8} & \textbf{58.5} \\
            \bottomrule[1pt]
        \end{tabular}
    \label{tab:abla_thought_order}
\end{table}

\begin{table}[h]
\centering
\caption{\textbf{Component Analysis of Self-Distillation Reward.} All three terms---the conditional indicator ($\mathbb{I}$), semantic similarity ($S_{sim}$), and conciseness score ($S_{concise}$)---are requisite for optimal performance.}
\small
\setlength{\tabcolsep}{5pt}
\begin{tabular}{ccc|cc|cc}
    \toprule[1pt]
        \multicolumn{3}{c|}{Components} & \multicolumn{2}{c|}{RefCOCOg} & \multicolumn{2}{c}{ReasonSeg} \\
        $\mathbb{I}$ & $S_{sim}$ & $S_{concise}$ & \#Tok $\downarrow$ & cIoU $\uparrow$ & \#Tok $\downarrow$ & cIoU $\uparrow$ \\
        \midrule
        \offmark & \offmark & \offmark & 159.8 & 71.3 & 111.9 & 54.4 \\
        \offmark & \onmark & \onmark & 67.8 & 70.1 & 75.2 & 54.6 \\
        \onmark & \offmark & \onmark & 46.2 & 68.6 & 43.6 & 57.8 \\
        \onmark & \onmark & \offmark & 78.9 & 70.7 & 74.2 & 56.3 \\
        \rowcolor{gray!10} \onmark & \onmark & \onmark & \textbf{34.0} & \textbf{71.8} & \textbf{38.8} & \textbf{58.5} \\
        \bottomrule[1pt]
    \end{tabular}
\label{tab:abla_reward_components}
\end{table}

\noindent\textbf{Inference-Time Prompting Strategies.}
Table~\ref{tab:abla_inference_time_prompt} evaluates how explicit instructions activate the learned reasoning policy. 
While the default WISE setting (simply omitting $\tau_d$) yields compression on ReasonSeg, it relies on the model implicitly inferring the stop condition. 
We observe that this implicit behavior is inherently unstable and inconsistent when applied across different contexts (as detailed in Sec.~\ref{sec:efficiency_analysis}), as the model lacks a definitive signal to terminate generation.
Adding our proposed brevity prompt (WISE-S, ``\textit{one sentence}'') resolves this ambiguity, serving as a robust activation signal that strictly enforces the learned concise policy.
In contrast, extreme prompts (``\textit{shorter the better}'') force over-compression, harming accuracy. Thus, WISE-S represents the optimal strategy to stabilize the model's concise reasoning capabilities.

\begin{table}[h]
\centering
\caption{\textbf{Effect of Inference Strategies.} While the default WISE setting ($+\tau_c-\tau_d$) offers substantial compression, the targeted brevity prompt in WISE-S achieves the optimal efficiency-accuracy balance.}
\small
\setlength{\tabcolsep}{4pt}
\begin{tabular}{l|cc|cc}
    \toprule[1pt]
        \multirow{2}{*}{Inference Strategy} & \multicolumn{2}{c|}{RefCOCOg} & \multicolumn{2}{c}{ReasonSeg} \\
        & \#Tok $\downarrow$ & cIoU $\uparrow$ & \#Tok $\downarrow$ & cIoU $\uparrow$ \\
        \midrule
        Seg-Zero (Baseline) & 159.8 & 71.3 & 111.9 & 54.4 \\
        WISE-F (+$\tau_c+\tau_d$) & 188.5 & 71.3 & 197.5 & 58.8 \\
        \midrule
        \multicolumn{5}{l}{\textit{Ablation: Prompt Modification}} \\
        $-\tau_c~~~~~~-\tau_d$ & 51.5 & 70.3 & 59.9 & 54.8 \\
        $-\tau_c~~~~~~+\tau_d$ & 200.0 & 68.4 & 221.4 & 53.2 \\
        WISE ($+\tau_c-\tau_d$) & 34.0 & \underline{71.8} & 38.8 & \textbf{58.5} \\
        \midrule
        \multicolumn{5}{l}{\textit{Ablation: Brevity Prompting}} \\
        \textit{``the shorter the better"} & \textbf{15.4} & 71.6 & \textbf{15.4} & 58.2 \\
        \rowcolor{gray!10} \textbf{WISE-S (\textit{``one sentence"})} & \underline{22.9} & \textbf{72.1} & \underline{22.7} & \underline{58.3} \\
        \bottomrule[1pt]
    \end{tabular}
\label{tab:abla_inference_time_prompt}
\end{table}

\begin{figure}[h]
    \centering
    \includegraphics[width=0.48\textwidth]{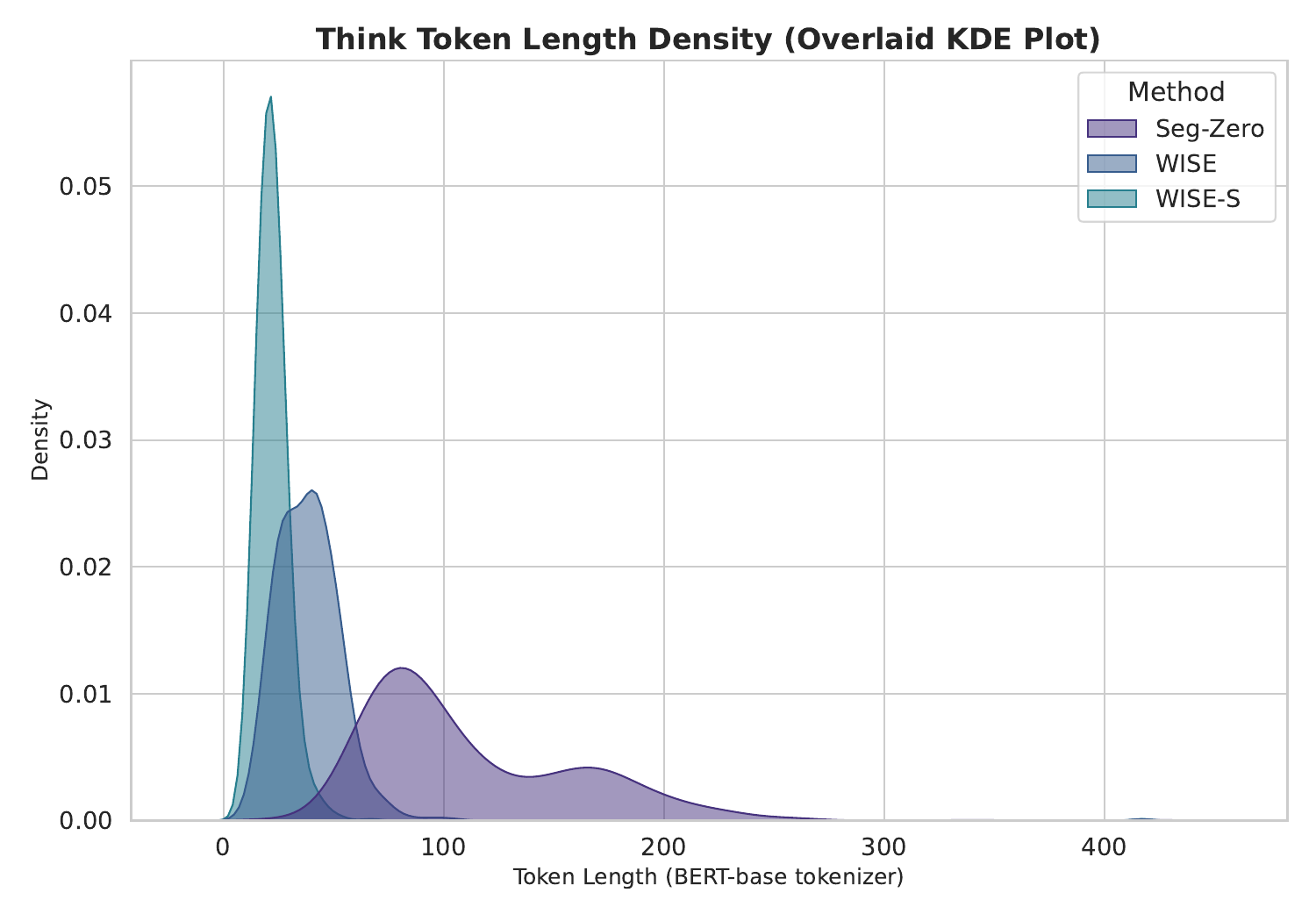} 
    \caption{\textbf{Distribution of Reasoning Token Length.} The KDE plot illustrates the dramatic reduction in both length and variance. WISE-S (green) demonstrates a tightly concentrated preference for conciseness compared to the long-tailed Seg-Zero baseline (purple).}
    \label{fig:token_distribution}
\end{figure}

\begin{figure*}[t]
    \centering
    \includegraphics[width=0.95\textwidth]{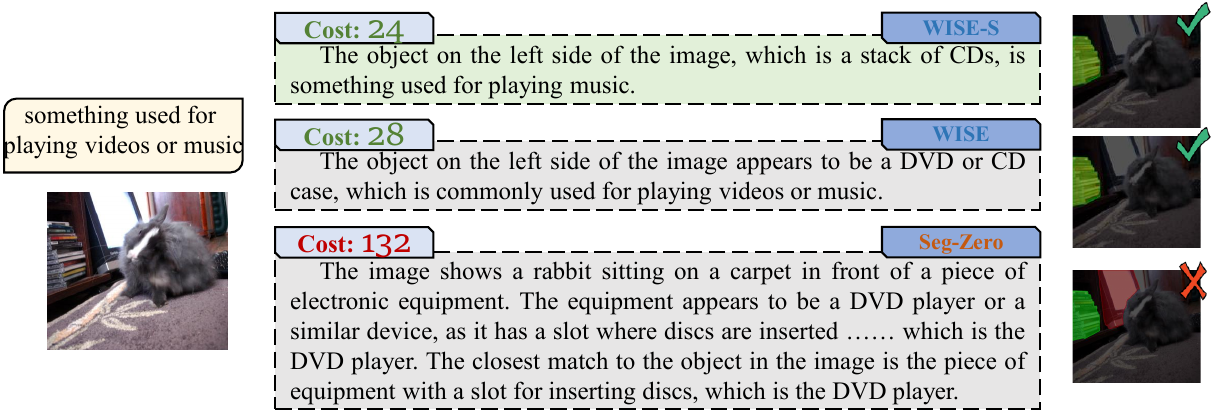} 
    \caption{\textbf{Qualitative Comparison.} Given a complex instruction, Seg-Zero produces a convoluted, 132-token rationale and fails. In contrast, WISE-S generates a concise, correct reasoning chain, successfully identifying the target with a fraction of the cost.}
    \label{fig:qualitative_comparison}
\end{figure*}

\subsection{Efficiency and Scalability Analysis}
\label{sec:efficiency_analysis}

Beyond methodological design, we analyze the system-level performance of WISE regarding wall-clock latency, token distribution, and scalability across architectures.

\noindent\textbf{Wall-Clock Efficiency and Baseline Validation.}
We quantify real-world gains in Table~\ref{tab:efficiency_baseline}. 
First, the failure of the raw Qwen2.5-VL (41.2 cIoU) validates our choice of the supervised Seg-Zero as a rigorous baseline.
Second, we investigate the efficacy of our inference prompting strategy largely independent of training. We apply our brevity prompt to the baseline (denoted as {Seg-Zero-S}). While this simple heuristic yields marginal gains, it fails to achieve the deep compression of our full framework (57.9 tokens vs. 22.7 for WISE-S). 
This highlights a dual contribution: 1) our prompting strategy is a transferable heuristic applicable to general models, but 2) the self-distillation training in WISE is the distinct driver for the {5$\times$ wall-clock speedup} (50.0m $\to$ 10.4m), which prompting alone cannot replicate.

\begin{table}[h]
    \centering
    \caption{\textbf{Wall-Clock Efficiency Analysis.} Comparison of inference latency and token costs. While the raw model underperforms and prompting alone (Seg-Zero-S) yields limited gains, WISE-S achieves a 5$\times$ speedup over the Seg-Zero baseline.}
    \label{tab:efficiency_baseline}
    \setlength{\tabcolsep}{3pt}
    \small
    \begin{tabular}{l|ccc|ccc}
        \toprule[1pt]
        \multirow{2}{*}{Method} & \multicolumn{3}{c|}{RefCOCO$_{\texttt{testA}}$} & \multicolumn{3}{c}{ReasonSeg$_{\texttt{test}}$} \\
        & {\#Tok} & {Time} & {cIoU} & {\#Tok} & {Time} & {cIoU} \\
        \midrule
        Qwen2.5-VL-7B & 0.0 & 11.2 & 77.8 & 0.0 & 7.6 & 41.2 \\
        Seg-Zero (Baseline) & 165.1 & 153.1 & 78.8 & 111.9 & 50.0 & 54.4 \\
        Seg-Zero-S (Prompt) & 85.6 & 116.1 & 79.0 & 57.9 & 20.3 & 55.0 \\
        \midrule
        \textbf{WISE (Ours)} & 35.0 & 28.9 & \textbf{79.1} & 38.8 & 17.0 & \textbf{58.5} \\
        \rowcolor{gray!10} \textbf{WISE-S (Ours)} & \textbf{24.0} & \textbf{16.1} & \textbf{79.1} & \textbf{22.7} & \textbf{10.4} & 58.3 \\
        \bottomrule[1pt]
    \end{tabular}
\end{table}

\noindent\textbf{Scalability across Architectures.}
To verify robustness, we extend WISE to different model scales (3B vs. 7B) and families (Qwen2 vs. Qwen2.5). As reported in Table~\ref{tab:scalability}, the method proves highly scalable. For the smaller Qwen2.5-VL-3B, WISE-S reduces token usage by over $4\times$ while slightly improving cIoU. Similar gains are observed on Qwen2-VL-7B, confirming that the thought compression mechanism is model-agnostic.

\begin{table}[h]
    \centering
    \caption{\textbf{Scalability across Architectures.} Performance consistency of WISE across different model families (Qwen2 vs. Qwen2.5) and parameter scales (3B vs. 7B).}
    \label{tab:scalability}
    \setlength{\tabcolsep}{3pt}
    \small
    \begin{tabular}{l|ccc|ccc}
        \toprule[1pt]
        \multirow{2}{*}{Method} & \multicolumn{3}{c|}{ReasonSeg$_{\texttt{val}}$} & \multicolumn{3}{c}{ReasonSeg$_{\texttt{test}}$} \\
        & {\#Tok} & {gIoU} & {cIoU} & {\#Tok} & {gIoU} & {cIoU} \\
        \midrule
        \multicolumn{7}{l}{\textit{Architecture: Qwen2-VL-7B}} \\
        Qwen2-VL-7B & 0.0 & 44.5 & - & 0.0 & 38.7 & - \\
        WISE (Ours) & 32.7 & 45.8 & 42.6 & 31.2 & 44.2 & 40.4 \\
        \rowcolor{gray!10} WISE-S (Ours) & 23.8 & 44.9 & 42.5 & 23.1 & 44.7 & 40.9 \\
        \midrule
        \multicolumn{7}{l}{\textit{Scale: Qwen2.5-VL-3B}} \\
        Seg-Zero-3B & 103.5 & 56.5 & 47.1 & 102.0 & 53.3 & 48.5 \\
        WISE-3B (Ours) & 34.5 & 57.0 & 52.8 & 33.4 & 53.8 & 48.6 \\
        \rowcolor{gray!10} WISE-3B-S (Ours) & 25.8 & 56.8 & 53.1 & 23.9 & 54.0 & 49.0 \\
        \bottomrule[1pt]
    \end{tabular}
\end{table}

\noindent\textbf{Distribution and Qualitative Analysis.}
To provide a more nuanced understanding of the efficiency gains, we visualize the density distribution of reasoning token lengths on the ReasonSeg test set in Figure~\ref{fig:token_distribution}. The baseline {Seg-Zero} exhibits a wide and long-tailed right-skewed distribution, indicating its reasoning process is not only long but also highly variable. In sharp contrast, our {WISE} model's distribution forms a sharp peak tightly concentrated around a much smaller mean, a result of its learned intrinsic preference for brevity. The {WISE-S} variant further narrows this distribution into a consistent, low-cost output, making its computational cost stable.
This efficiency gain is not merely a reduction in length but a reflection of improved reasoning quality. As illustrated in the qualitative example in Figure~\ref{fig:qualitative_comparison}, given a complex functional instruction, Seg-Zero engages in a convoluted, 132-token rationale and ultimately fails. Conversely, WISE-S identifies the correct object with a focused, 24-token thought. This vividly demonstrates that our thought compression encourages the model to discard irrelevant details and focus on the core logic, leading to reasoning that is not only more efficient but also more robust. More examples are provided in the Appendix~\ref{sec:qualitative_analysis}.

\section{Conclusion} \label{sec:conclusion}

We introduced \textbf{WISE} to resolve the tension between reasoning depth and efficiency in LMMs via thought compression. By enforcing a predictive ``concise-first'' generation order and utilizing conditional self-distillation, our method learns reasoning abstraction without human annotations. Experiments confirm that WISE, especially the WISE-S variant, achieves state-of-the-art zero-shot performance on ReasonSeg while reducing computational costs by over 4$\times$. This proves that deep reasoning can be efficient, enabling practical real-world applications.

\section*{Impact Statement}

This work contributes to the advancement of efficient and sustainable multimodal AI. By reducing reasoning token usage by over 5$\times$, WISE lowers the energy consumption per inference, aligning with Green AI initiatives to mitigate the environmental footprint of large-scale model deployment. Furthermore, our self-distillation framework demonstrates that effective reasoning policies can be learned without relying on expensive human-annotated rationales, offering a data-efficient pathway for scaling reasoning capabilities. Finally, the significant reduction in latency enhances the feasibility of integrating complex reasoning into real-time, high-throughput applications, helping bridge the gap between academic research and practical production systems.

\nocite{langley00}

\bibliography{ref}
\bibliographystyle{icml2026}

\clearpage
\newpage
\appendix

\section{Qualitative Analysis: Distilling Success and Failure}
\label{sec:qualitative_analysis}

To validate the robustness of our thought compression, we analyze both success and failure scenarios. Our analysis reveals that WISE-S achieves high efficiency without introducing new reasoning errors, maintaining high \textbf{Semantic Fidelity} to the original verbose reasoning.

\begin{figure*}[t] % 使用 figure* 可以让图跨双栏显示 (如果是双栏论文)，单栏论文用 figure 即可
    \centering
    
    % 第一个子图
    \begin{subfigure}[b]{0.95\linewidth}
        \centering
        \includegraphics[width=\linewidth]{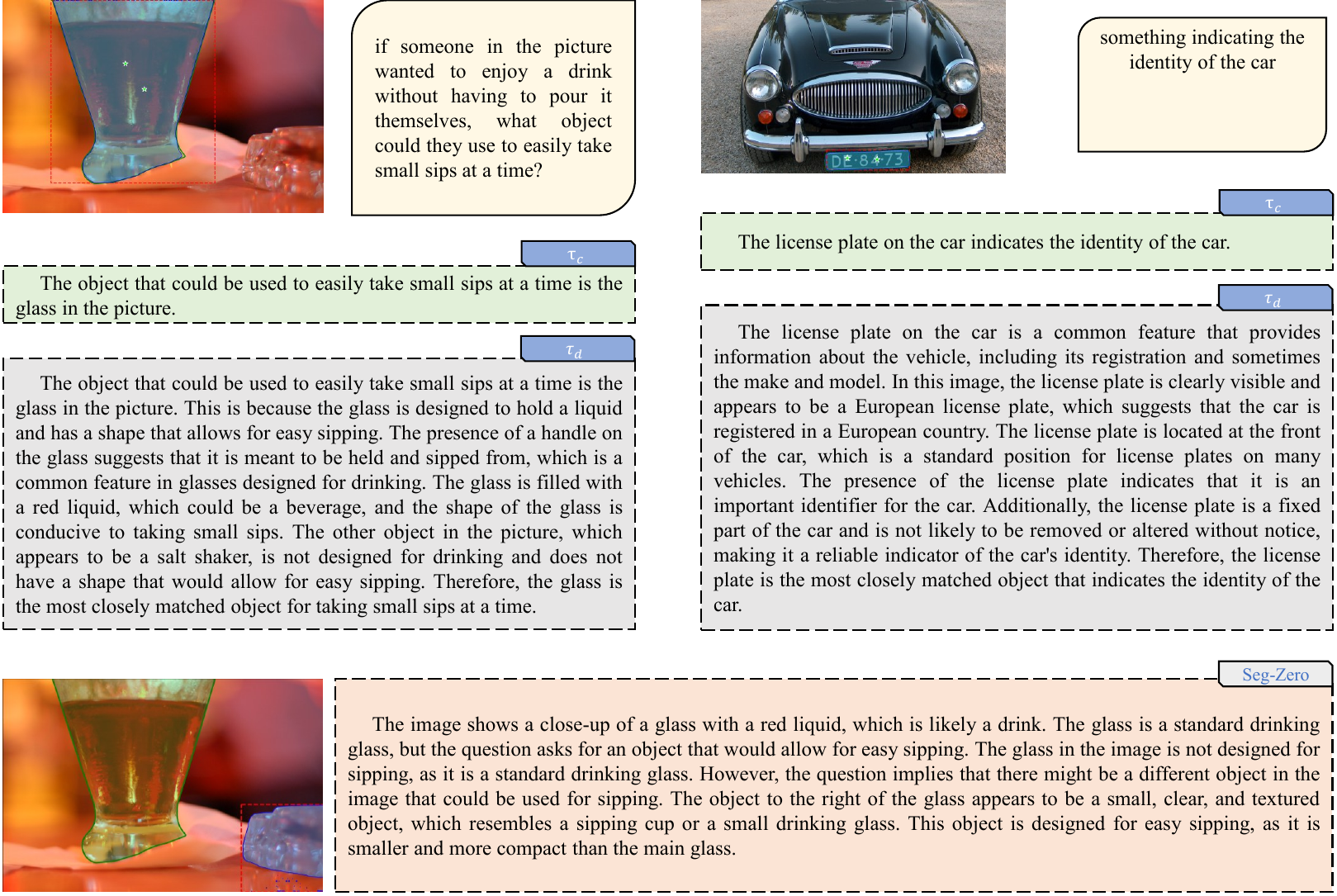}
      %   \caption{Functional Reasoning Case} % 如果不需要(a)标题，可以留空或注释掉这一行
         \caption{\textbf{Qualitative Comparison: Efficiency vs. Distraction.} 
The figure demonstrates how WISE-S acts as a sufficient statistic, contrasting its focused reasoning with the baseline's verbose failure. 
\textbf{(Left) Functional Reasoning:} While WISE-S ($\tau_c$) efficiently identifies the glass by its affordance (``take small sips''), the baseline \textbf{Seg-Zero} (bottom) suffers from \textit{reasoning drift}. It over-analyzes the prompt constraints, leading to a hallucinated conclusion about a non-existent ``small textured object'' to the right. This vivid example illustrates how thought compression can prevent the model from getting lost in irrelevant visual details.
\textbf{(Right) Identity Grounding:} WISE-S correctly filters out spatial redundancy to focus on the discriminative visual attribute (``license plate'') required to establish identity.}
        \label{fig:good_case_1}
    \end{subfigure}
    
    \vspace{1em} % 调整两图之间的垂直间距
    
    % 第二个子图
    \begin{subfigure}[b]{0.95\linewidth}
        \centering
        \includegraphics[width=\linewidth]{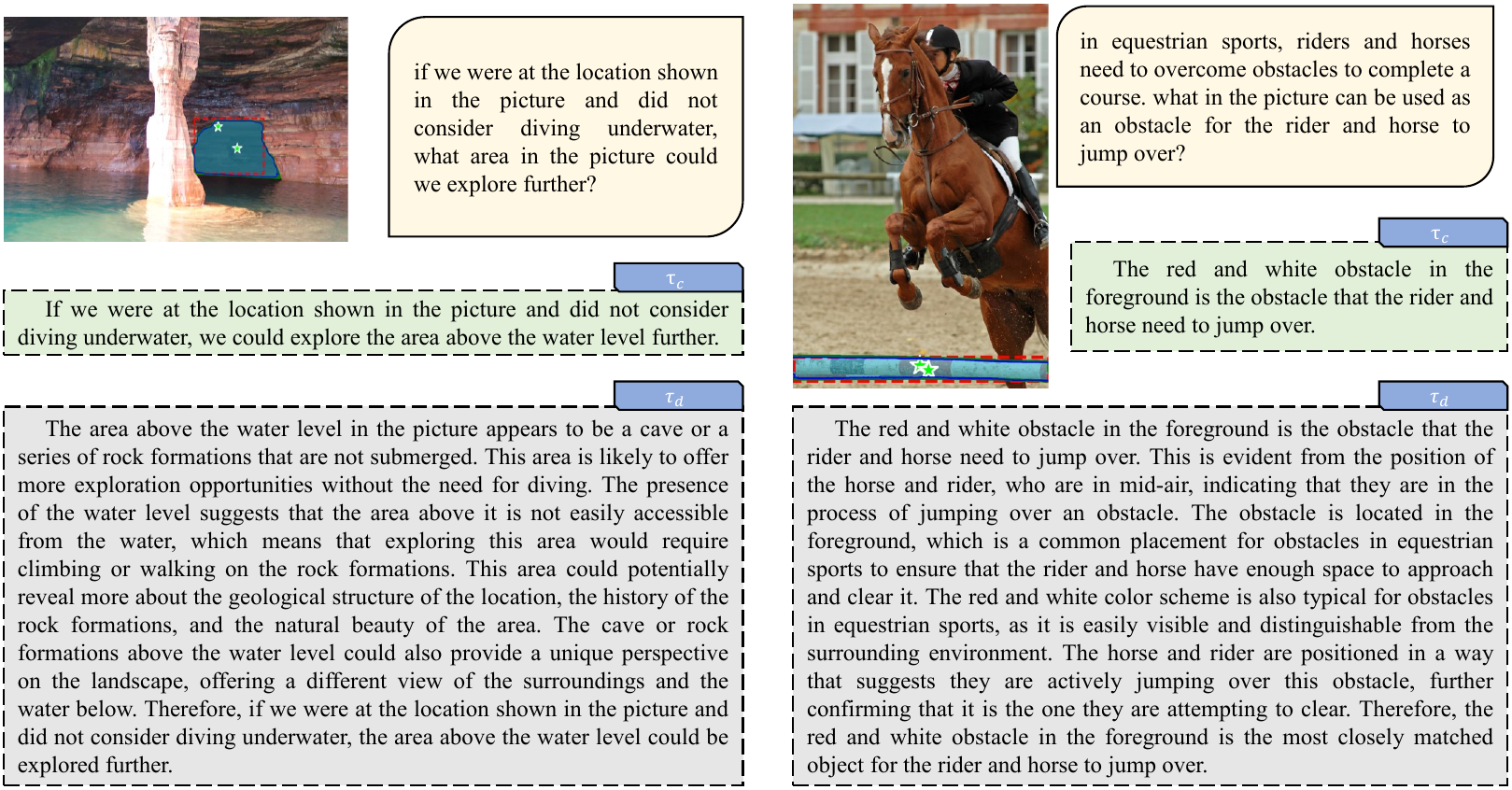}
        \caption{Visual Attribute Grounding Case} % 如果不需要(b)标题，可以留空或注释掉这一行
        \label{fig:good_case_2}
    \end{subfigure}
    
    \caption{\textbf{Success Case Study Examples.} The figures demonstrate how WISE-S acts as a sufficient statistic. (a) Shows the model efficiently identifying an object by its function (``take small sips'') without verbose description. (b) Shows the model correctly filtering out spatial redundancy to focus on discriminative visual attributes (``red and white obstacle'').}
   %  \caption{}
    \label{fig:good_cases}
\end{figure*}

\textbf{Success Cases: Concise Rationale as a Sufficient Summary.}
As shown in Figure~\ref{fig:good_cases}, in successful instances, WISE-S demonstrates the ability to distill the \textit{decision logic} while discarding \textit{visual redundancy}.
\begin{itemize}
    \item \textit{Attribute Grounding:} In the equestrian example (Figure~\ref{fig:good_cases}, Bottom-Right), the detailed explanation ($\tau_d$) engages in a verbose verification process, checking the horse's position and the nature of the sport. In contrast, WISE-S ($\tau_c$) directly extracts the discriminative features—``red and white obstacle'' and ``foreground''—which are sufficient to localize the mask.
    \item \textit{Functional Reasoning:} In the cave exploration example (Figure~\ref{fig:good_cases}, Bottom-Left), the model must process a negative constraint (``did not consider diving''). WISE-S correctly reasons that the target area must be ``above the water level,'' efficiently pruning the search space without the need for the extensive geological description found in the detailed chain.
\end{itemize}
These examples confirm our hypothesis that the concise rationale learns to act as a sufficient summary for the final answer, effectively bridging the gap between instruction and segmentation.

\begin{figure*}[t]
    \centering
    
    % 第一个坏案例子图
    \begin{subfigure}[b]{0.95\linewidth}
        \centering
        \includegraphics[width=\linewidth]{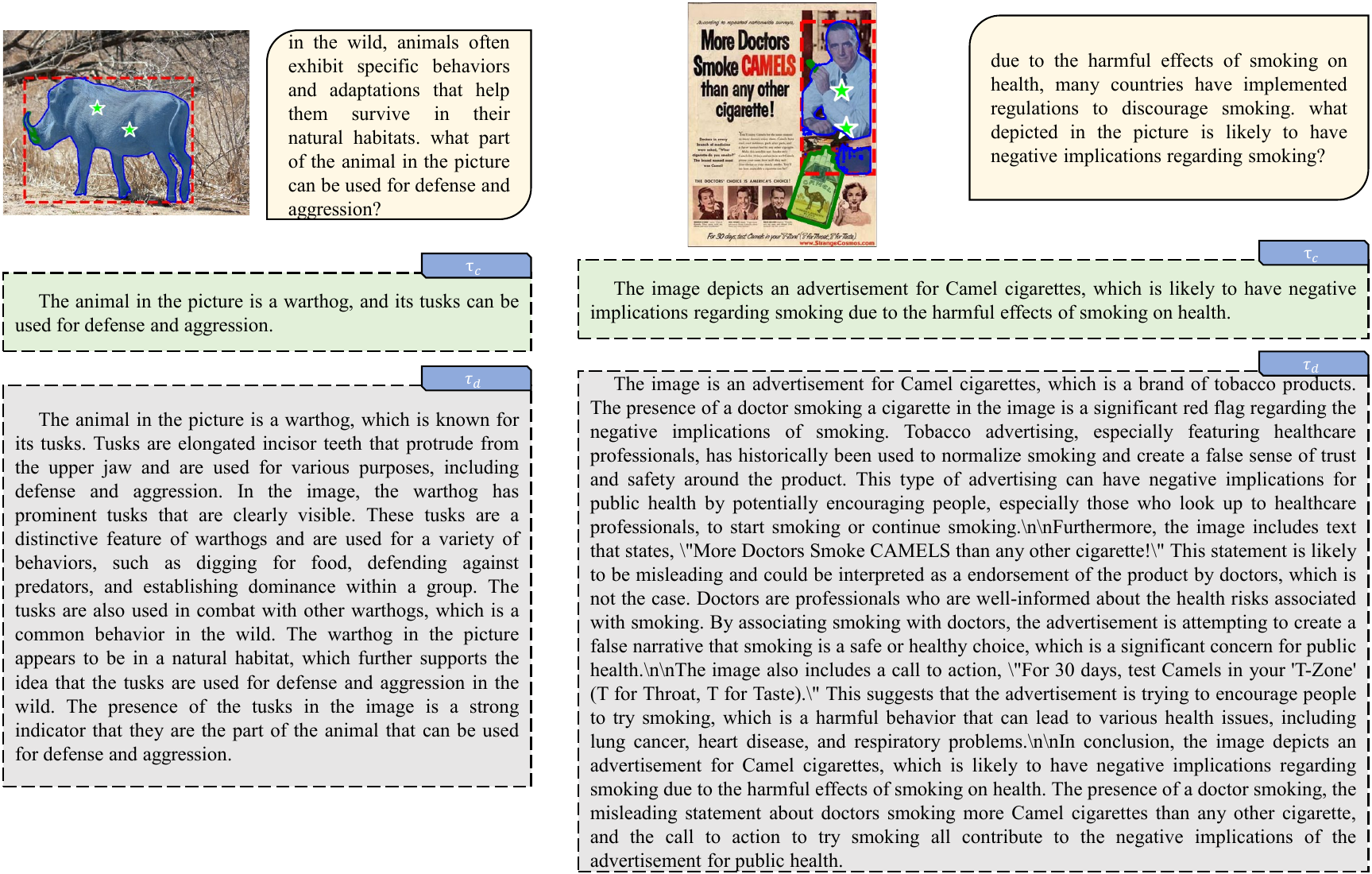}
        \caption{Grounding Limitation Case} 
        \label{fig:bad_case_1}
    \end{subfigure}
    
    \vspace{1em} % 调整两图之间的垂直间距
    
    % 第二个坏案例子图
    \begin{subfigure}[b]{0.95\linewidth}
        \centering
        \includegraphics[width=\linewidth]{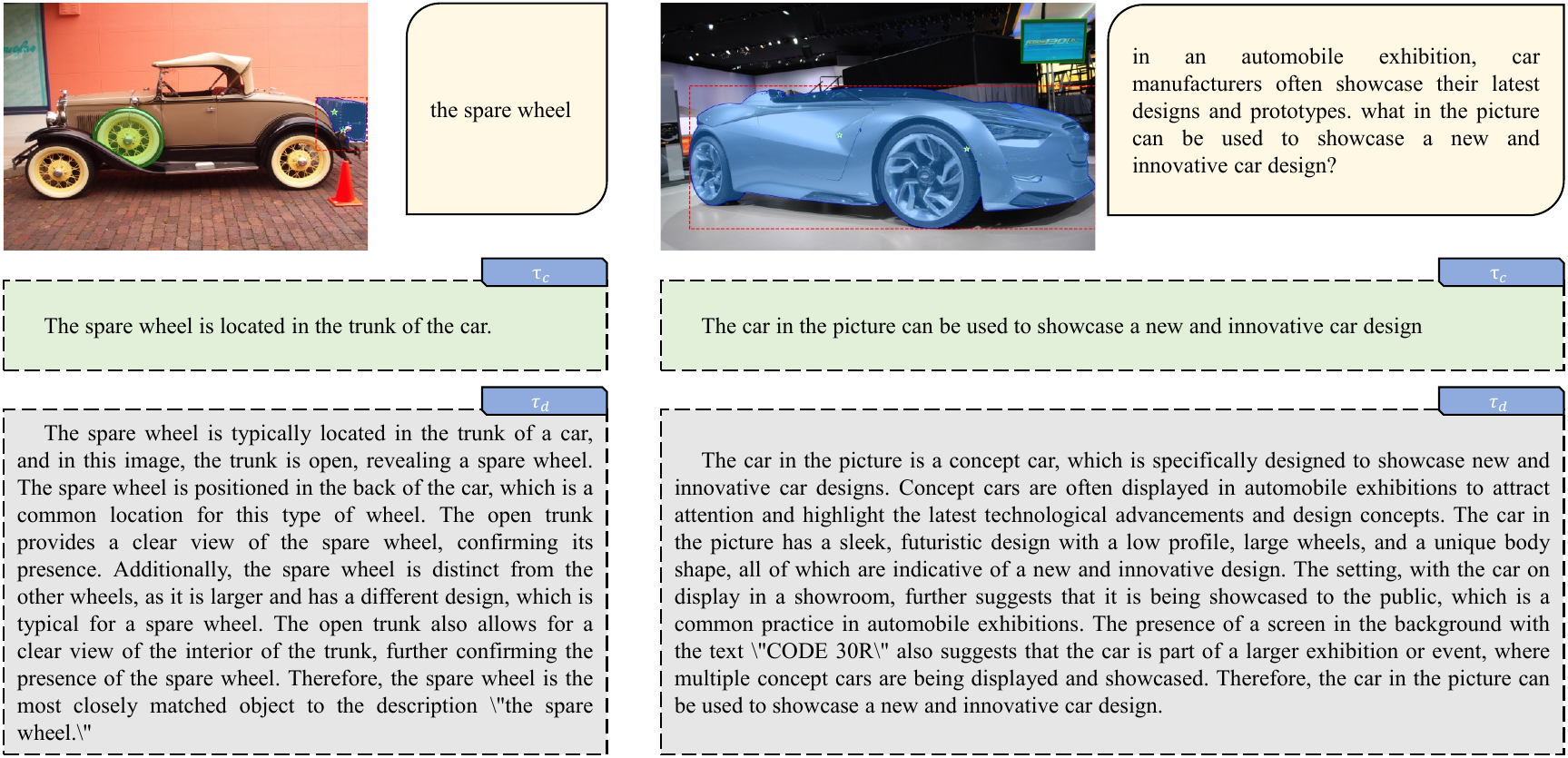}
        \caption{Circular Logic Case} 
        \label{fig:bad_case_2}
    \end{subfigure}
    
    \caption{\textbf{Analysis of Failure Cases: Semantic Fidelity amidst Grounding Errors.} 
    These examples illustrate instances where the model fails to produce an accurate mask (Low IoU). Crucially, however, the Concise Rationale ($\tau_c$) remains a \textbf{faithful summary} of the Detailed Explanation ($\tau_d$). 
    (a) In the top example, both $\tau_c$ and $\tau_d$ correctly identify the semantic topic but struggle to ground the abstract concept to specific pixels. 
    (b) In the bottom example, both rationales exhibit circular logic without identifying distinct visual attributes.
    This indicates that the failures stem from the \textbf{underlying limitations} of the base model's spatial grounding capabilities or reasoning loops, rather than information loss caused by the WISE-S compression mechanism.}
    \label{fig:bad_cases}
\end{figure*}

\textbf{Failure Cases: Consistent Limitations.}
We closely examined cases with low IoU (Figure \ref{fig:bad_cases}) to determine if the brevity constraint caused the failure. Interestingly, we found that \textbf{the compression mechanism is rarely the culprit}.
\begin{itemize}
    \item \textbf{Fidelity in Failure:} In the ``Warthog'' example, the concise rationale ($\tau_c$) correctly identifies the target object as ``tusks,'' fully capturing the semantic core of the verbose explanation ($\tau_d$). The failure to segment the specific tusks (likely masking the whole face) is a shared limitation in \textit{spatial grounding} inherent to the base Vision-Language Model, occurring in both standard WISE and WISE-S modes.
    \item \textbf{Shared Hallucination/Ambiguity:} In the ``Concept Car'' example, both $\tau_c$ and $\tau_d$ exhibit circular logic (tautology), failing to identify specific visual attributes. $\tau_c$ merely summarizes the vague reasoning of $\tau_d$.
\end{itemize}
\textbf{Conclusion:} These failure cases powerfully demonstrate the effectiveness of our \textbf{Self-Distillation} objective. The model successfully internalized the reasoning—whether strong or weak—into a compressed form. The errors stem from the backbone model's capabilities, not the thought compression process itself. This confirms that WISE-S provides a ``lossless'' speedup in terms of reasoning quality.

\end{document}